\documentclass[10pt,twocolumn,letterpaper]{article}

\usepackage{cvpr}
\usepackage{times}
\usepackage{epsfig}
\usepackage{graphicx}
\usepackage{amsmath}
\usepackage{amssymb}

\usepackage{etoolbox}
\makeatletter
\patchcmd{\@makecaption}
  {\scshape}
  {}
  {}
  {}
\makeatother

\usepackage{bm}
\usepackage{xcolor}
\usepackage{graphics} 
\usepackage{epsfig} 
\usepackage{mathptmx} 
\usepackage{times} 
\usepackage{amsmath} 
\usepackage{amssymb}  
\usepackage{graphicx}
\usepackage{multirow, makecell}
\usepackage{array}
\usepackage{booktabs} 
\usepackage{comment}
\usepackage{commath}
\usepackage{subfig}
\usepackage{cite}
\usepackage[linesnumbered,ruled]{algorithm2e} 

\usepackage{ragged2e}
\newcolumntype{P}[1]{>{\RaggedRight\hspace{0pt}}p{#1}}
\newcolumntype{M}[1]{>{\centering\arraybackslash}m{#1}}

\newcommand{\beginsupplement}{%
        \setcounter{table}{0}
        \renewcommand{\thetable}{S\arabic{table}}%
        \setcounter{figure}{0}
        \renewcommand{\thefigure}{S\arabic{figure}}%
        \setcounter{section}{0}
        \renewcommand\thesection{\Alph{section}}
     }

\usepackage[pagebackref=true,breaklinks=true,letterpaper=true,colorlinks,bookmarks=false]{hyperref}

\cvprfinalcopy 

\def\cvprPaperID{10268} 

\ifcvprfinal\fi
\begin{document}

\title{CVAE-H: Conditionalizing Variational Autoencoders via Hypernetworks and Trajectory Forecasting for Autonomous Driving}

\author{Geunseob (GS) Oh, Huei Peng\\
University of Michigan\\
{\tt\small gsoh@umich.edu, hpeng@umich.edu}
}

\maketitle
\thispagestyle{empty}

\begin{abstract}
The task of predicting stochastic behaviors of road agents in diverse environments is a challenging problem for autonomous driving. To best understand scene contexts and produce diverse possible future states of the road agents adaptively in different environments, a prediction model should be probabilistic, multi-modal, context-driven, and general. We present Conditionalizing Variational AutoEncoders via Hypernetworks (CVAE-H); a conditional VAE that extensively leverages hypernetwork and performs generative tasks for high-dimensional problems like the prediction task. We first evaluate CVAE-H on simple generative experiments to show that CVAE-H is probabilistic, multi-modal, context-driven, and general. Then, we demonstrate that the proposed model effectively solves a self-driving prediction problem by producing accurate predictions of road agents in various environments. 
\end{abstract}

\section{Introduction}
\label{sec:introduction}

Over the last few years, companies have made major progress towards the real-world deployment of autonomous driving technologies. In late 2020, Waymo, Google's autonomous vehicle (AV) company, became the first company that offered commercial autonomous taxi rides to the public in Pheonix, Arizona. With the initial milestone for the deployment made, the challenge of developing effective AV technologies that can operate in a wide variety of complex urban environments has never been more important. As pointed out by a recent publication from Waymo, generalization within and between operating regions is crucial to the overall viability of AV technologies \cite{waymo_open_dataset_scalability_2020}. 

Autonomous vehicles interact with stochastic road users such as human drivers in different driving scenarios and road topologies. The presence of multiple stochastic agents creates uncertainties that grow over time, contributing to the increase in the number of variations of the environments and scenarios that an autonomy stack (perception, prediction, planning, and control) of the AV needs to handle. Among the modules of the autonomy stack, the prediction is where the stochasticity and diversity culminates, as it is responsible of understanding scene contexts of various complex environments and predicting a set of possible future trajectories of road users.

\begin{figure}[t] 
    \centering
    \includegraphics[width=0.85\linewidth]{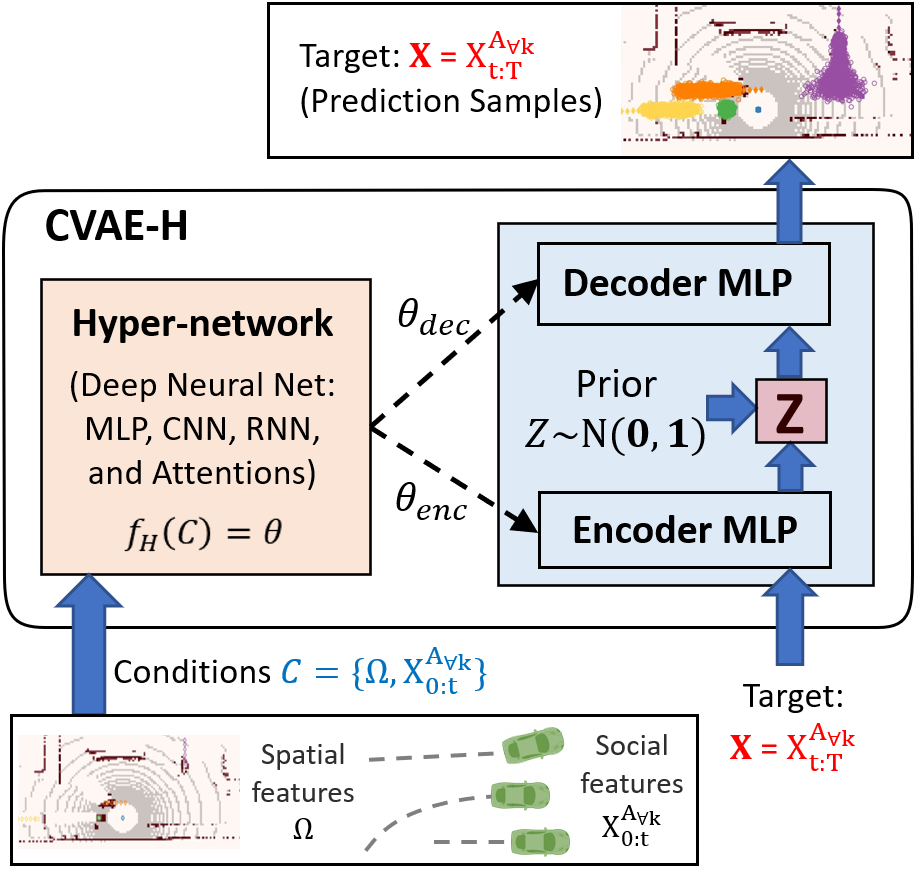}
    \caption{CVAE-H is a conditional VAE that integrates a hypernetwork into a VAE. In a trajectory forecasting problem, CVAE-H takes a high-dimensional features (e.g., lidar and labelled trajectories) as conditions and produces diverse predictions for dynamic road agents in the scene.}
    \label{Fig:front_page_figure}
\end{figure}

There are four key attributes that a prediction module should possess: \emph{probabilistic}, \emph{multi-modal}, \emph{context-driven}, and \emph{general}. Autonomous driving often involves interactions with stochastic road users. Upon the presence of road users, AV needs to predict their states to avoid collisions. In addition to bare predictions, a prediction model should also output the quantified uncertainties of the predictions, which are essential in the downstream modules such as planning.

Another characteristic of road agents is the diversity of possible trajectories. For example, a vehicle approaching a 4-way intersection has four possible \emph{modes}; go straight, turn left, turn right, or yield if there is a lead vehicle waiting in the queue. A vehicle passing through a crosswalk may slow down or maintain its speed. 

There are various factors that contribute to the diversity of the trajectories. Such factors include the relative states of dynamic agents, road geometries, and static obstacles in the scene. To consider these factors, it is important to capture the impact of both social features (e.g., interactions among road-agents) and spatial features (e.g., lanes, stop signs, and crossroads) in the prediction.

Last attribute of autonomous driving is that numerous instances of driving environments exist. Road configuration alone is diverse; examples include un-signalized intersections with single-lane, two-lanes, 2-way stops, or u-turns, one-way roads, straight roads, curved roads, and so on. The presence of static and dynamic agents diversify the urban environments as well. This means that an ad-hoc prediction approach customized to a specific instance of an environment is impractical. A scalable prediction model should adaptively produce predictions in various environments. 



It is not trivial for a prediction model to satisfy all four attributes. First of all, not all neural-network models are probabilistic unless they build on explicit density models such as Mixture Density Network (e.g., Gaussian Mixture Models) \cite{bishop1994MDN, oh2019pred_TL}, Normalizing-flow models \cite{rhinehart2018r2p2, oh2020hcnaf}, Variational Autoencoder (VAE) \cite{lee2017desire}. Secondly, not all probabilistic models are multi-modal, for instance, a uni-modal Gaussian. Third, to be fully context-driven, the model should leverage both the social and spatial information. Some models \cite{casas2018intentnet, luo2018fast_and_furious} are based exclusively on Convolutional Neural Network (CNN) and may not best capture social features. Similarly, models that rely solely on social features \cite{alahi2016socialLSTM, gupta2018socialGAN, oh2020STL_falsification} may not consider spatial features of the environment that come from HD Maps, images, and/or lidar point clouds. Finally, generality of a prediction model is hard to achieve. Models that leverage scenario-specific heuristics or knowledge may sacrifice generality over the precision for the targeted scenarios. 

To build a model that possesses all four attributes of a ideal prediction model, we leverage the following two architectures: VAE \cite{kingma2013VAE}, which is an explicit generative model, and hypernetwork \cite{ha2016hypernetworks}, which may consist of diverse kinds of neural networks for the context-driven prediction. By integrating a hypernetwork into a VAE, we propose CVAE-H. This facilitates the informational flow of the conditions to the VAE network across all layers and allows the VAE to effectively scale up to work with large conditional inputs. We first evaluate CVAE-H on simple Gaussian experiments to show that CVAE-H is probabilistic, multi-modal, context-driven, and general. Then, we demonstrate that the proposed model effectively solves a self-driving prediction problem by utilizing both social and spatial features of the environments and producing accurate predictions of road agents in various environments.

\section{Related Works}
\label{sec:pred_lit_review}

In this section, we review existing models for the prediction task and categorize them. The first categorization of the prediction approaches is based on input features the prediction model utilizes. Models that focus on the social features include Social-LSTM \cite{alahi2016socialLSTM}, Social-GAN \cite{gupta2018socialGAN}, and Trafficpredict \cite{ma2019trafficpredict}. Some models focused on exploiting the spatial features, for example, IntentNet \cite{casas2018intentnet}, Fast and Furious \cite{luo2018fast_and_furious}, that effectively capture underlying traffic rules from the road geometry; however, interactions among road agents are not explicitly modeled. There are also models that utilize both social and spatial features such as DESIRE \cite{lee2017desire}, SoPhie \cite{ref:2019_Sophie}, R2P2 \cite{rhinehart2018r2p2}, and MTP \cite{tang2019MFP}. 

The second categorization is based on the core mechanics of the prediction models; either deterministic or probabilistic. Many earlier deterministic models leveraged either physics-based modeling techniques \cite{treiber2000IDM} (e.g., constant velocity model and car-following model) or relatively light-weight neural networks such as multi-layer perceptron (MLP), RNN, and/or CNN \cite{alahi2016socialLSTM}, while a few recent deterministic models work with larger deep neural networks \cite{casas2018intentnet, luo2018fast_and_furious, tang2019MFP}. Generative models are examples of probabilistic models that model a joint distribution $p(x_{i})$ using $z$ that is a variable that encodes the underlying dynamics of the data distribution. Furthermore, a group of generative models that are used to model a conditional distribution $p(x_{o}|x_{i})$ are called \emph{conditional generative models}. Examples include conditional VAE \cite{ref:2015_cVAE}, conditional GAN \cite{ref:2014_cGAN}, and conditional flow \cite{oh2020hcnaf}.

Modern generative models, including GAN, VAE, Flow, and Mixture Density Networks (MDN) \cite{bishop1994MDN} are popular across several domains of artificial intelligence and machine learning due to their capacities to approximate complex probability distributions. GAN is well-known for producing samples of excellent qualities. However, it does not explicitly model the probability \cite{goodfellow2016GANtutorial}. This means that GAN has to rely on expensive Monte-Carlo sampling to approximate the probability distribution of the generated samples to reason the uncertainties of the predictions. 

Unlike GAN, VAE is a type of \textit{explicit} generative model that models probability density $p(x)$ explicitly. This property allows us to obtain the uncertainties of the sample predictions directly. Besides, several recent VAE-based models have shown promising results at predicting trajectories of human-driven vehicles in urban environments \cite{lee2017desire, tang2019MFP, ivanovic2020pred_CVAE}. 

Normalizing flow, or Flow, is also a type of explicit generative model. While flow-based models \cite{rhinehart2018r2p2, rhinehart2019precog, oh2020hcnaf} have access to the exact probability density $p(x)$, flow-based models are typically better suited for density estimations than generative tasks (e.g., trajectory forecasting).

MDN is another generative model that models explicit probability densities. A common choice for the density model is a mixture of Gaussians where the neural network estimates the parameters of the Gaussian mixture. MDN with Gaussian mixtures have been used for prediction tasks \cite{oh2019pred_TL, ma2019trafficpredict, chai2019multipath}. Since MDN works with a pre-determined number of modes, a MDN(GMM)-based model may sacrifice the generality.

\section{Preliminaries}
\label{sec:prelim}

In this section, we review the deep learning concepts that are the two main keys of CVAE-H; Variational autoencoder \cite{kingma2013VAE} and Hypernetwork \cite{ha2016hypernetworks}.

\subsection{Variational AutoEncoder (VAE)}
\label{sec:prelim_VAE}

Variational Autoencoder \cite{kingma2013VAE} is a generative model that learns a data distribution $p(X)$ in an unsupervised way by using an abstract (i.e., latent) variable $Z$ that captures the underlying dynamics of the data distribution $X$.

As the name suggest, VAE is an \textit{autoencoder} model \cite{hinton2006AE} trained using variational inference \cite{jordan1999VI}. Autoencoder is an unsupervised model that learns an identity function using an encoder $f_{enc}$ and decoder $f_{dec}$. The encoder $f_{enc}(X)=Z$ takes the input data $X$ that are high dimensional and compresses it into a lower dimensional data $Z$. The decoder $f_{dec}(Z)=X$ then takes $Z$ and recovers the higher dimensional input $X$. During the reconstruction of the original inputs, the autoencoder aims to discover a more efficient lower dimensional representation of the inputs. 

VAE combines the autoencoder with variational inference. Instead of mapping the input into a fixed vector, VAE maps the input into probability distributions; the encoder and decoder each models $p(Z|X)$ and $p(X|Z)$, which are both typically parameterized using neural networks. As the compression $X \rightarrow Z$ and generation $Z \rightarrow X$ steps become probabilistic, VAE can be used for generative tasks by sampling $Z$ and passing it through the decoder to obtain $X$. 

Let us denote $p(Z|X)$ and $p_{m}(Z|X)$ as the true posterior and model posterior distributions. Likewise, $p(X|Z)$ and $p_{m}(X|Z)$ are denoted as the true likelihood and model likelihood. Finally, we denote $p(Z)$ as the prior distribution. The learning objective is to maximize the likelihood of the training data to find the optimal parameters $\theta^{*} = argmax(\prod^{N}_{k=1}p_{m}(X_{k}))$, where $p(X_{k}) = \int_{Z} p(X_{k}|Z)p(Z)dZ$ via the use of law of total probability. 

The computation of $p(X_{k})$ is intractable, but can be approximated using variational inference \cite{jordan1999VI}. Instead of maximizing the likelihood, a computable term called \textit{Evidence Lower BOund (ELBO)} of the likelihood can be maximized. As the name indicates, ELBO is a lower bound of the original likelihood, i.e., $p(X) \geq ELBO$. Precisely, the following relationship holds: $log(p(X)) = ELBO + KL(p_{m}(Z|X_{k})||p(Z|X_{k}))$, where ELBO is defined as $\mathbb{E}_{Z}\Big[log(p(X_{k}|Z))\Big] - KL(p_{m}(Z|X_{k})||p(Z))$. The ELBO is easily computable as long as the posterior, prior, and generative distributions are modeled using explicit density models and through the use of reparameterization trick \cite{kingma2013VAE}.

VAE approximates the target distribution and is typically used to generate new data or estimate the probability density of the target approximately. In practice, VAE-based models have demonstrated competitive results for trajectory prediction tasks \cite{lee2017desire, ivanovic2020pred_CVAE} as well as other machine learning tasks such as sequence modeling and density estimation \cite{chung2015recurrent_vae, kingma2016IAFVAE, higgins2016beta}.

\subsection{Hypernetwork}
\label{sec:prelim_HN}

\begin{figure}[ht] 
    \centering
    \includegraphics[width=0.8\linewidth]{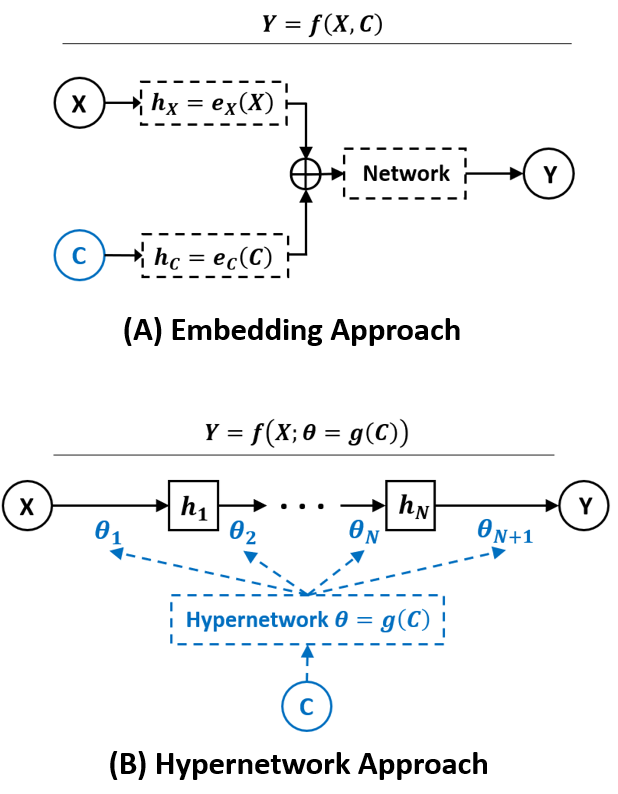}
    \caption{Two approaches to build conditional neural nets. $X, Y, C$ each indicates input, output, and conditions.}
    \label{Fig:embedding_vs_hypernetwork}
\end{figure}

Assume a function that takes two inputs $X, C$ and outputs $Y$, i.e., $f:(X, C) \rightarrow Y$. Given that $f$ is a probabilistic mapping, $p(Y|X,C)$ represents a probability distribution of the output $Y$ conditioned on the two inputs. Learning such $f$ and the corresponding conditional distribution can be achieved using two different approaches; (1) the embedding and (2) the hypernetwork approach, as depicted in Figure \ref{Fig:embedding_vs_hypernetwork}.

As commonly used in machine learning, the embedding approach is a straightforward way of learning $f$. This approach implies a model that consists of the embedding modules $e_{1}(X) = h(X)$ and $e_{2}(C) = h(C)$ where the two outputs are concatenated $h := h(X) \oplus h(C)$ and passed to another network $f(h) = Y$.  

The hypernetwork approach \cite{ha2016hypernetworks} is a less intuitive solution that relies on a hierarchical structure between two models; one called a primary network $g$ produces weights of the other separate network $f$. This corresponds to $f(X;\theta)=Y$ where $\theta = g(C)$. Observe that the embedding approach uses an embedding network $e$ and treats $C$ equally to $X$ as part of the input. On the other hand, the hypernetwork approach models the conditional mapping $f_{C}(X)=Y$, i.e., learning the contributions of $X$ to $Y$ via a parameterized function $f(X;\theta)$ where the parameters themselves are the outputs of another network $g$, which is independent of $f$.

The ability to effectively learn conditional functions (i.e., model a function that transforms into different functions depending on the condition $C$) corresponding to the property of modularity \cite{galanti2020HN_modularity} is a major benefit of the hypernetwork approach. A recent study showed that hypernetwork exhibits the property of modularity for a large $g$ and small $f$ and the hypernetwork approach uses much smaller networks compared to the embedding approach to achieve the same performance in modeling conditional distributions \cite{galanti2020HN_modularity}. Furthermore, hypernetworks offer excellent \textit{generalization} capability when it is combined with deep neural networks. This has been demonstrated across diverse tasks in machine learning \cite{oh2020hcnaf, bertinetto2016HN_oneshot, suarez2017HN_highway, brock2018HN_smash, zhang2018HN_graph_NAS, von2019HN_continual, lorraine2018HN_stochastic_optim, galanti2020HN_modularity}.

Hypernetwork provides a way to make the unconditional generative models (e.g., vanilla VAE and normalizing flow) conditional without a use of RNN. Hypernetwork is a neural-network that can be modeled using various types of neural networks modules such as CNN, RNN, Residual blocks \cite{ref:residual_connection_he2016}, and Attention \cite{ref:attention_vaswani2017}. The flexibility of the hypernetwork allows the generative models to even condition on bigger models such ResNet \cite{ref:resnet_he2016}, BERT \cite{ref:devlin2018bert}, or FPN \cite{lin2017feature} by including such models as part of the hypernetwork. In this sense, hypernetworks take diverse types of data including social and spatial data. The effectiveness of the hypernetworks grows with respect to the capacity of its backbone neural network. Upon use of CNN blocks (spatial features) and RNN/Attention modules (social features), the main neural network can become \textit{context-driven}.

This becomes especially useful for the prediction task for autonomous driving. A challenge of the prediction task is that it may involve very large inputs whose dimensions are in the magnitude of millions. This is because the inputs include social features (e.g., labeled states for all road-agents in the vicinity of the AV) and spatial features (e.g., lidar scans, camera images, and/or HD maps) from past to the current moment. Since our prediction model extensively leverages hypernetworks, the high-dimensional information about the driving environment can be effectively encoded and used for contextual prediction.

\section{CVAE-H}
\label{sec:pred_CVAE-H}

We propose CVAE-H (Conditionalized Variational AutoEncoder via Hypernetwork), a conditional VAE that integrates a hyper-network into a VAE. The hypernetwork encodes various conditioning information (e.g., social and spatial information) and the VAE utilizes the encoding as the network parameters to perform auto-encoding $X>Z>X$ and generative $Z>X$ tasks.

As elaborated in Section \ref{sec:prelim_HN}, one advantage of hypernetwork over the embedding approach is the flexibility of hypernetwork (i.e., any differentiable neural network can be used as a hypernetwork). In addition, the hypernetwork passes the information from the conditions to the VAE network across all VAE layers and thus allows the VAE to scale up to large conditional inputs of the prediction problem. Another advantage is that hyper-network can encode conditioning information with a smaller number of network parameters compared to the embedding approach \cite{galanti2020HN_modularity}.

\begin{figure}[ht] 
    \centering
    \includegraphics[width=\linewidth]{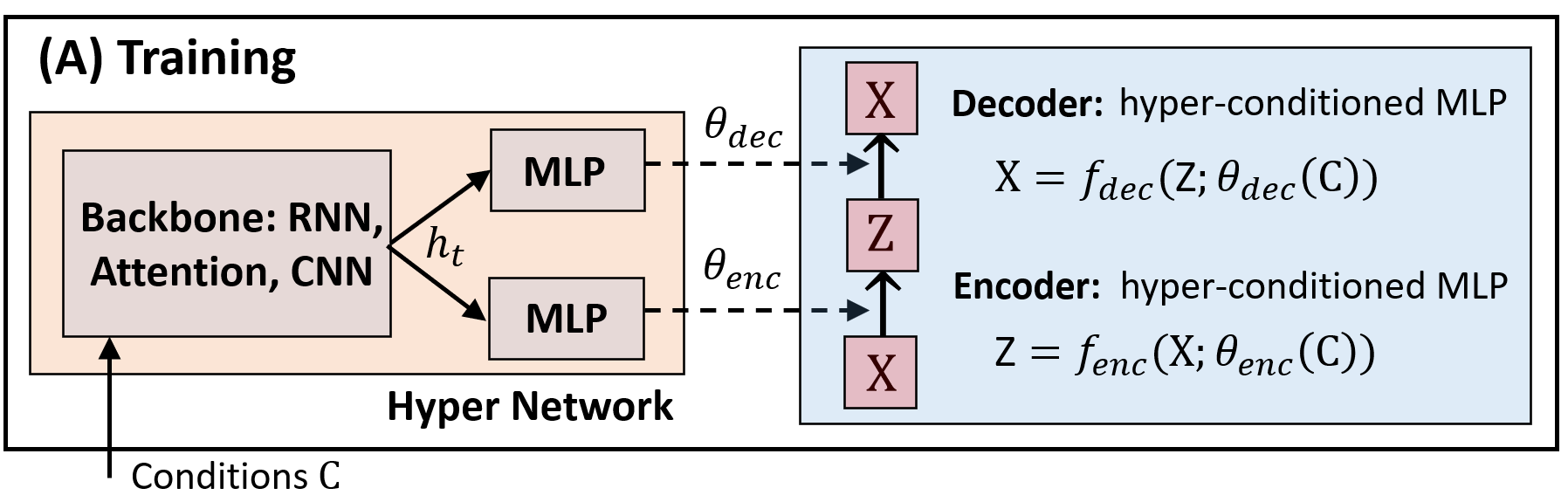}
    \caption{Schematics of the CVAE-H architecture. During the training, CVAE-H takes conditions $C$ and a target variable $X$ and learns the encoder and decoder to reconstruct $X$. For the prediction problem, we use the following conditions $C:=[X^{A_{\forall k}}_{-t:0}, \Omega]$ and target variable $X_{0:T}:=[x^{A_{\forall k}}_{0:T}, y^{A_{\forall k}}_{0:T}]$.}
    \label{Fig:CVAE-H_schematic_training}
\end{figure}

Figure \ref{Fig:CVAE-H_schematic_training} illustrates CVAE-H, which largely consists of two modules: 1) a VAE and 2) a hypernetwork that computes the network parameters of 1). In the following subsections, the details of the two modules are presented.

\subsection{Hyper-conditioning VAE}

Conditionalizing a VAE via a hypernetwork involves the following process of parameterizing the encoder and decoder of the VAE:

\vspace{-5pt}
\begin{align}
\label{eq:cond_KL}
(Encoder) \quad Z &= f_{enc}(X; \theta_{enc}(C)) \nonumber
\\
(Decoder) \quad X &= f_{dec}(Z; \theta_{dec}(C))
\end{align}

where $\theta_{enc}$ and $\theta_{dec}$ are obtained via the hypernetwork $f_{H}(C)$ as in $f_{H}(C) = [\theta_{enc}, \theta_{enc}]$. If a VAE is conditionalized using the embedding approach, the resulting equations would instead be $Z=f_{enc}(X,g_{1}(C))$ and $X=f_{dec}(Z,g_{2}(C))$. For the reasons described in Section \ref{sec:prelim_HN}, we conditionalize the VAE using a hypernetwork. 

The VAE architecture can consist any type of neural network. For a 1D target variable such as a sequence of coordinates or a collection of variables, MLP is a good choice. While we expect CNN to be an effective encoder \& decoder structure for a 2D target variable such as an image, we leave the search of proper network for 2D target variables to the future work.

Similar to how $p(X)$ is modeled in an unconditional VAE, a latent variable $Z$ is introduced to map the inputs $X$ and condition $C$ to the the conditional distribution $p(X|C)$ as follows: 
$p(X|C)=\int_{Z} p(X|Z,C)p(Z|C)dZ$. Here, $p(Z|C)$ and $p(X|Z,C)$ represent the prior and the generative decoder conditioned on $C$, respectively. Ideally, $Z$ should encode abstract features of $p(X|C)$.

Since the integral to compute $p(X|C)$ is intractable, variational inference \cite{jordan1999VI} is used to approximate the integral with a parameterized posterior $q_{\theta_{enc}}(Z|X,C)$. This leads to the construction of ELBO on the log conditional likelihood:

\begin{align}
\label{eq:ELBO_cVAE}
\log p(X|C) \geq \mathbb{E}_{q_{\theta_{enc}}(Z|X,C)}[\log p_{\theta_{dec}}(X|Z,C)] \nonumber \\
 - \mathrm{KL}(q_{\theta_{enc}}(Z|X,C)||p(Z|C)).
\end{align}

We set the conditional prior $P(Z|C)$ to be a zero-mean standard multi-variate Gaussian (i.e., diagonal covariance with all entries equal to one). Different choices of the prior include parameterizing priors using neural networks and/or domain knowledge \cite{oord2017VQVAE}.

\subsection{Hypernetwork and Training of CVAE-H}

The hypernetwork aims to effectively propagate the conditioning information to the VAE. An earlier work that extensively leveraged hypernetwork to conditionalize a normalizing-flow \cite{oh2020hcnaf} showed that the main-network becomes versatile by customizing the backbone neural networks of the hypernetwork. In this regard, for sequential targets, a hypernetwork can include RNN, attention modules, and/or Transformer \cite{ref:attention_vaswani2017}. For computer vision tasks, a hypernetwork can use CNN-based models. 

The goal of CVAE-H is to learn the target conditional distribution $p(X|C)$ by minimizing the negative log-likelihood of model distribution $p_{model}(X|C)$. This is achieved by maximizing the conditional ELBO presented in Equation \ref{eq:ELBO_cVAE} (i.e., the ELBO is the loss function). The first term of the ELBO is the reconstruction error of the VAE and can be computed as long as the estimation of $p_{\theta_{dec}}(X|Z,C)$ is tractable. For this reason, we design the decoder to be an explicit density model (e.g., softmax, MDN \cite{bishop1994MDN}).
In this work, we model the decoder to output mixture parameters for a Gaussian mixture: the mean $\mu$, covariance $\Sigma$, and mixing probabilities of the output Gaussian mixture $\rho$. The second term of the ELBO is a KL divergence and can be easily computed as we use a multi-variate standard Gaussian for the prior $p(Z|C)$ and a multi-variate (uni-modal) Gaussian with a diagonal covariance for the posterior $q_{\theta_{enc}(Z|X,C)}$.


\subsection{Inference: CVAE-H for Generative Tasks}
\label{sec:CVAE-H_generative_task}

\begin{figure}[ht] 
    \centering
    \includegraphics[width=\linewidth]{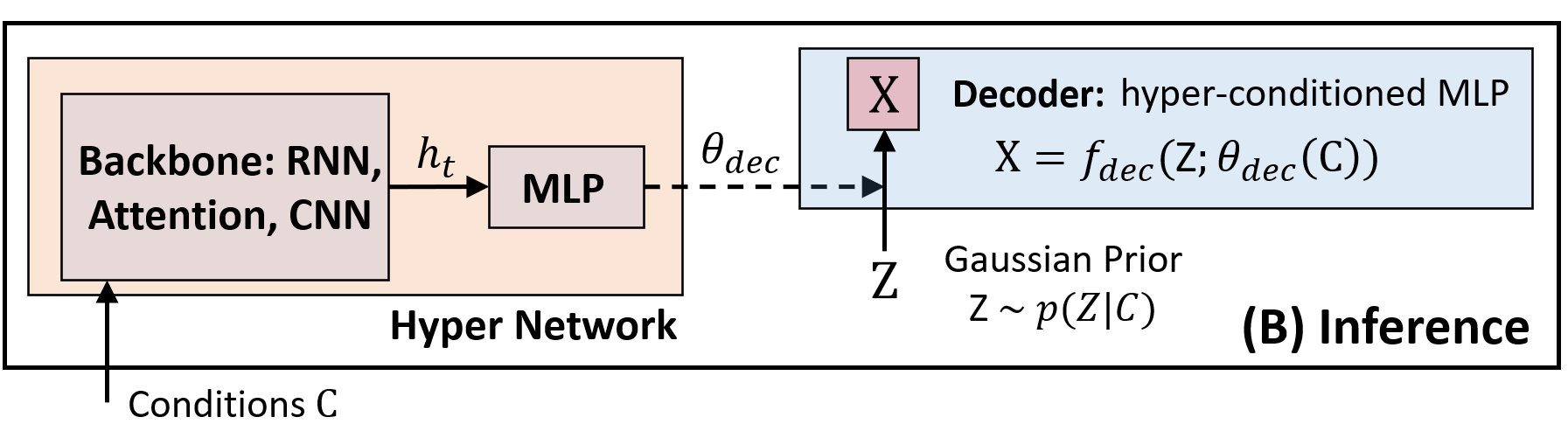}
    \caption{A schematic of the CVAE-H during inference. As opposed to the training, CVAE-H does not take $X$ and only utilizes the conditions $C$. Instead, $Z$ is sampled using the prior and is passed through the decoder to obtain $X$.}
    \label{Fig:CVAE-H_schematic_inference}
\end{figure}

The training and inference processes for CVAE-H are different. During the training, we have access to the target variable $X$ and we run the encoder $Z = f_{enc}(X;\theta_{enc}(C))$ to obtain the posterior $Z$. The decoder then takes $Z$ to recover $X = f_{dec}(Z;\theta_{dec}(C))$. As illustrated in Figure \ref{Fig:CVAE-H_schematic_inference}, during the inference, the encoder is not used because we do not have access to $X$. Instead, we leverage the Gaussian prior distribution $p(Z|C)$, sample latent variable $Z \sim p(Z|C)$, and run the decoder to generate $X = f_{dec}(Z;\theta_{dec}(C))$. Similar to the training, $C$ is passed to the hyper-network and $\theta_{dec}$ is computed to construct the decoder $f_{dec}$, but $\theta_{enc}$ is not computed as the we do not use the encoder in the inference. In this regard, CVAE-H can perform both the \emph{auto-encoding} of a target variable like a vanilla VAE ($X > Z > X$) and \emph{generative process} ($Z > X$) via the inference process described above. We note that we use the generative process for the self-driving prediction problem as we do not have access to the target variable $X$ during the inference.

\section{Experiments: Gaussian Generative Tasks}
\label{sec:gaussians}

The next two Sections are dedicated to the experiments we conducted to evaluate CVAE-H. This Section introduces the \emph{Gaussians} experiments that are designed to examine the effectiveness of CVAE-H on generative tasks. Then in the next Section, we evaluate CVAE-H on a prediction task for autonomous driving, which is a more challenging problem that involves high-dimensional spatio-temporal sensor information such as lidar scans of the environments and labelled trajectories. 

In this Section, we assess the model in simple generative tasks used in the literature \cite{papamakarios2017MAF, huang2018NAF, kingma2016IAFVAE, oh2020hcnaf}. The task of forecasting a distribution of trajectories is fundamentally a generative task. Therefore, generative tasks help us evaluate the quality of samples generated from CVAE-H.

The Gaussian experiments allow us to evaluate CVAE-H in simpler and more interpretable settings than the self-driving prediction task. They provide us with better insights regarding the strengths and weaknesses of CVAE-H. Most importantly, the experiments examine the capacity of CVAE-H as an ideal prediction model introduced in Section \ref{sec:introduction}. In other words, we validate if the proposed prediction model is probabilistic, multi-modal, context-driven, and general.

Two generative experiments, namely \textit{Gaussian 1} and \textit{Gaussian 2}, are used to verify whether CVAE-H have the four attributes of the ideal prediction model. To evaluate the performance of the model, we utilize both quantitative and qualitative measures. The quantitative measures include NLL (i.e., cross-entropy) and KL divergence ($D_{KL}$), which are defined in the following.

\begin{equation}
\begin{split}
\label{eq:def_NLL_KL}
& \textrm{NLL} \hspace{47pt}    = -E_{X\sim p(X|C)}[\textrm{log}(p_{model}(X|C))], \\
& D_{KL}(p || p_{model}) = \sum_{X\sim p(X|C)}p(X|C)\textrm{log}\bigg(\frac{p(X|C)}{p_{model}(X|C)}\bigg).
\end{split}
\end{equation}

To evaluate CVAE-H qualitatively, we visualize the outputs of the models. Recall that CVAE-H outputs sample coordinates $X \sim p(X|C)$. In this regard, the visualizations include the resulting probability densities of the generated samples from CVAE-H. Specifically, we discretize a 2D grid and estimate the density of each cell by counting the number of generated samples in the cell.

\begin{figure}[h] 
    \centering
    \includegraphics[width=\linewidth]{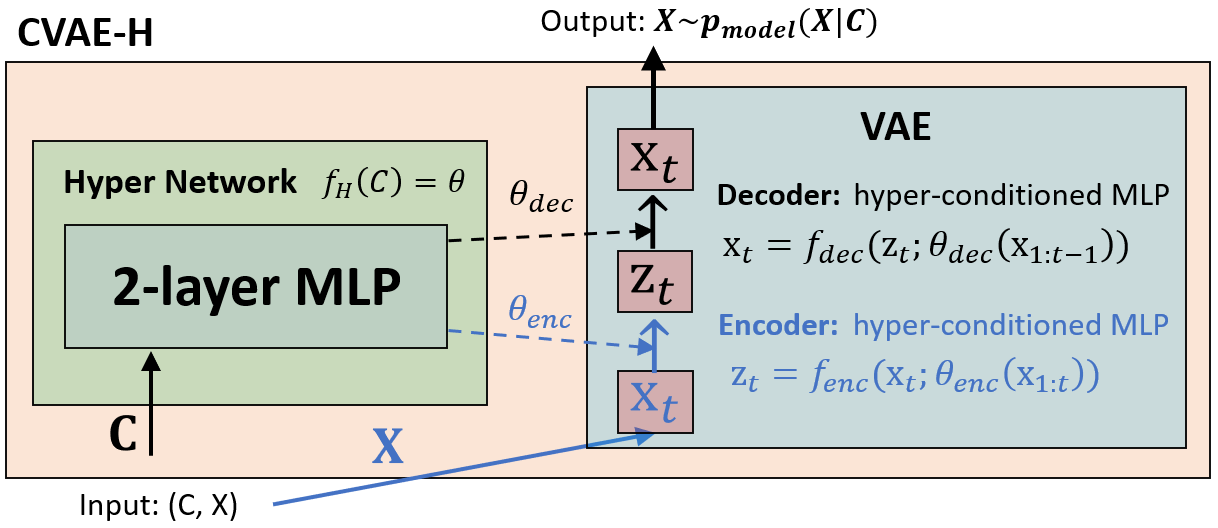}
    \caption{The CVAE-H customized for the Gaussian experiments. The inputs to the CVAE-H are a tuple of target variable $X$ and conditioning information $C$ where $X$ corresponds to the $(x,y)$ coordinates and $C$ corresponds to a set of discrete class variable $C \in \{0,1,2\}$ for the Gaussian 1 experiment and $C \in \{(0,0),(-4,4),(-4,4),(4,-4),(4,4)\}$ for the Gaussian 2 experiment. The parts colored blue are only used for the training. In the inference, Z is sampled from a bivariate standard Gaussian prior.}
    \label{Fig:pred_hypernet_gaussian_exp}
\end{figure}

Figure \ref{Fig:pred_hypernet_gaussian_exp} portrays the CVAE-H network designs customized for the Gaussian experiments. Note that a hypernetwork is a versatile neural network that aims to extract features from the conditioning information $C$. In this regard, a hypernetwork design changes depending on the task and types of $C$.

\subsection{The Gaussian 1 Experiment}

\begin{figure}[ht] 
    \centering
    \includegraphics[width=0.95\linewidth]{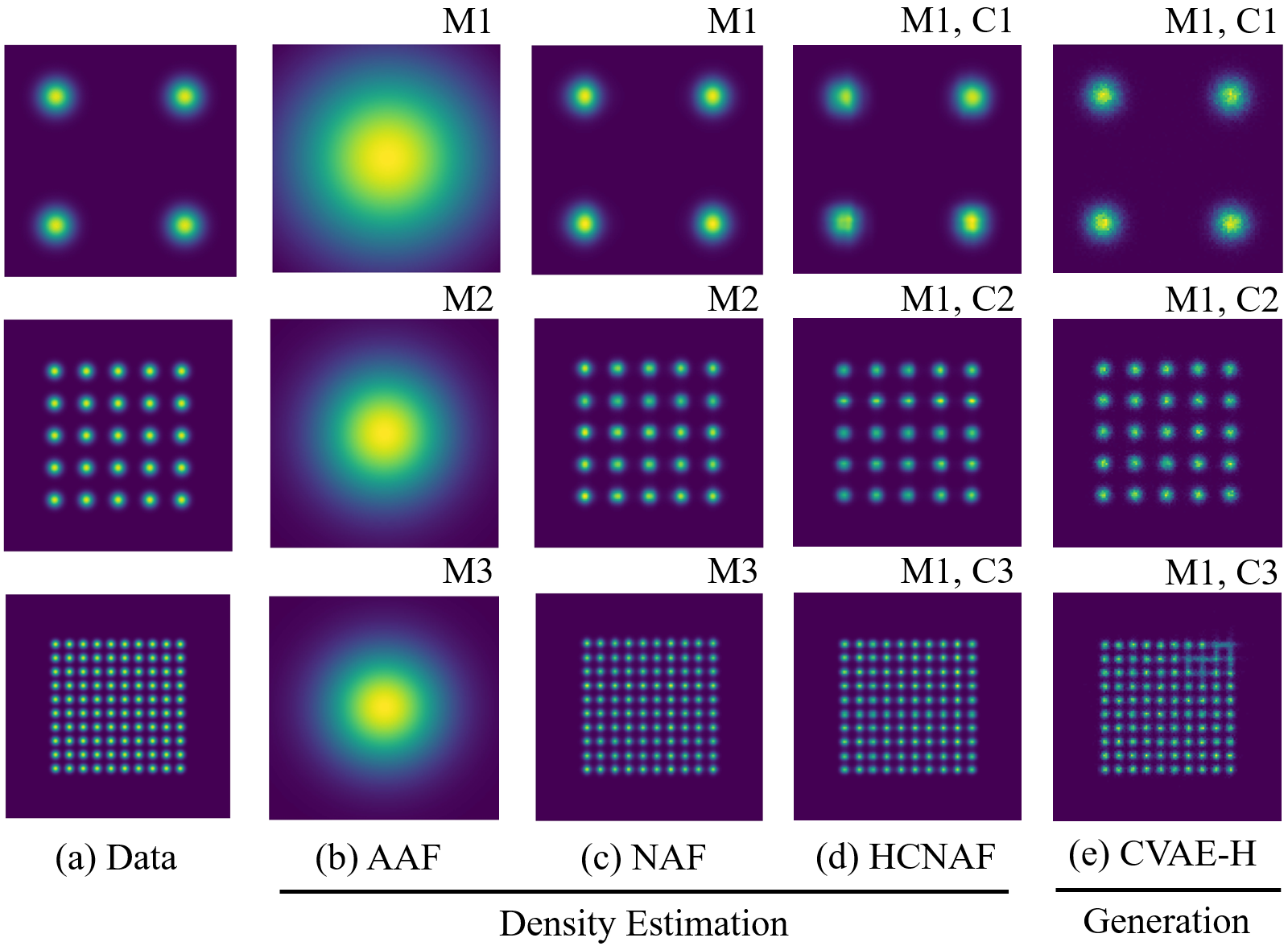}
    \caption{Qualitative result of the Gaussian 1 experiment. M and C each denotes model and condition respectably. This results suggest that CVAE-H models conditional distributions effectively, comparable to the HCNAF results.}
    \label{Fig:Gaussian_1_exp}
\end{figure}

The Gaussian 1 experiment is the experiment used in \cite{huang2018NAF, oh2020hcnaf} and aims to show the model's conditioning ability for three distinct multi-modal distributions $p_{1}(x,y), p_{2}(x,y), p_{3}(x,y)$ of 2-by-2, 5-by-5, and 10-by-10 Gaussians depicted in Figure \ref{Fig:Gaussian_1_exp}. We compare the results against a conditional generative model, HCNAF \cite{oh2020hcnaf}, which accurately reproduces conditional distributions using a single model.

In this experiment, the hypernetwork is designed using a 2-layer MLP with Relu activation functions and takes a condition $C \in \mathbb{R}^1$ and outputs $\theta_{enc}$ and $\theta_{dec}$. We design the VAE of CVAE-H using 4-layer hyper-conditioned MLPs with Relu activation functions for both encoder and decoder. The last layer of the decoder outputs parameters of a Gaussian mixture (i.e., means, variances, and mixing probabilities). The outputs of CVAE-H are generated by sampling the output Gaussian mixture. All the other parameters were set identically to the setting reported in \cite{oh2020hcnaf}, including those for the Adam optimizer (the learning rate $5e^{-3}$ decays by a factor of 0.5 every 2,000 iterations with no improvement in validation samples). The NLL (i.e., $p_{model}(X_{target})$) values in Table \ref{Table:Exp_Gaussian1} were computed using 10,000 samples.

In order to reproduce the three distributions, AAF and NAF require three different and separately trained models $p_{1}(x,y; \theta_{1}), p_{2}(x,y; \theta_{2}), p_{3}(x,y; \theta_{3})$. Conversely, a \textit{single} model of HCNAF ($p(x,y|C_{k};\theta)$) and CVAE-H ($p(x,y|Z,C_{k};\theta)$) can model all three conditions using a single set of network parameters $\theta$. The conditioning information $C_{k} \in \{0,1,2\}$ each value represents the class of the 2-by-2, 5-by-5, and 10-by-10 Gaussians. 

\begin{table}[h]
\centering
\caption{NLL for the experiment depicted in Figure \ref{Fig:Gaussian_1_exp}. Lower values are better.  Other than CVAE-H, we reuse the numbers reported in \cite{oh2020hcnaf}.}
\label{Table:Exp_Gaussian1}
\begin{tabular}{
>{\centering\arraybackslash}m{1.3cm}
>{\centering\arraybackslash}m{0.75cm} 
>{\centering\arraybackslash}m{0.75cm}
>{\centering\arraybackslash}m{1.0cm} 
>{\centering\arraybackslash}m{2.25cm} }
\toprule
 & AAF & NAF & HCNAF & CVAE-H (ours) \\
\toprule
2 by 2   & 6.056 & 3.775 & 3.896 & 3.895 \\
\midrule
5 by 5   & 5.289 & 3.865 & 3.966 & 3.978 \\
\midrule
10 by 10 & 5.087 & 4.176 & 4.278 & 4.292 \\
\bottomrule
\end{tabular}
\end{table}

Results from Figure \ref{Fig:Gaussian_1_exp} and Table \ref{Table:Exp_Gaussian1} show that CVAE-H is able to reproduce the three nonlinear target distributions, confirming that it are \emph{probabilistic} and \emph{multi-modal}. Depending on the condition $C_{k}$, CVAE-H produced three very different distributions, thus \emph{context-driven} with respect to simple conditions. CVAE-H also achieves comparable results to HCNAF, which is a powerful conditional density estimator that can model complex conditional distributions. It is important to note that our model uses a \emph{single} model to produce the 3 distinct pdfs like HCNAF, whereas AAF and NAF used \emph{3 distinctly trained} models. 

We point out that the plots displayed in Figure \ref{Fig:Gaussian_1_exp} contains results from both density estimation (for AAF, NAF, and HCNAF) and sample generation (for CVAE-H). In the density estimation task, a model evaluates $p_{model}(x,y|C_{k})$ where $(x,y)$ belongs to a cell in a discrete 2-dimensional grid. In other words, $(x,y) \in$ Grid$([x_{min},y_{min}], [x_{max},y_{max}])$. In the sample generation task, a set of outputs $(x,y) \sim p_{model}(x,y|Z,C_{k}; \theta), k=1,2,3$ are generated by passing $Z$, which is sampled from the prior, through the decoder. Then, a histogram is used to classify the generated $(x,y)$ into the 2-dimensional grid. While NLLs for the three density estimators and CVAE-H in the same table, the NLLs were evaluated differently. Precisely, the NLLs for the density estimators were estimated by evaluating $p_{model}(X|C_{k})$ whereas the NLL for CVAE-H was obtained by computing $p_{model}(X|Z,C_{k}), Z \sim N(0,I)$.

\subsection{The Gaussian 2 Experiment}

The Gaussian 1 experiment tested the model's ability to learn target distributions. On the other hand, the Gaussian 2 experiment is designed to evaluate how CVAE-H can generalize over \emph{unseen} conditions $C_{unseen}$. By assessing the model with a set of conditions beyond what it was trained with, we can evaluate the model' capacity to interpolate and extrapolate, which is important for the application of CVAE-H on problems such as a self-driving prediction task.

We train a single CVAE-H model to learn five distinct pdfs, where each pdf represents a Gaussian distribution with its mean used as conditions $C:=(x_{c}, y_{c}) \in \mathbb{R}^{2}$ and an isotropic standard deviation $\sigma$ of 0.5. We train the models with five different discrete conditions $C_{train} = \{C_{1}, ... , C_{5}\}$, where $C_{i}$ represents the mean of an isotropic bivariate Gaussian pdf. We then check how accurately the model (1) reproduces the data distribution $p(x,y|C_{train})$ it was trained on and (2) predict new distributions $p_{model}(x,y|C_{unseen})$, $C_{unseen}:=\{C_{6}, ... , C_{9}\}$ that the model did not see before.

\begin{figure}[h] 
    \centering
    \includegraphics[width=\linewidth]{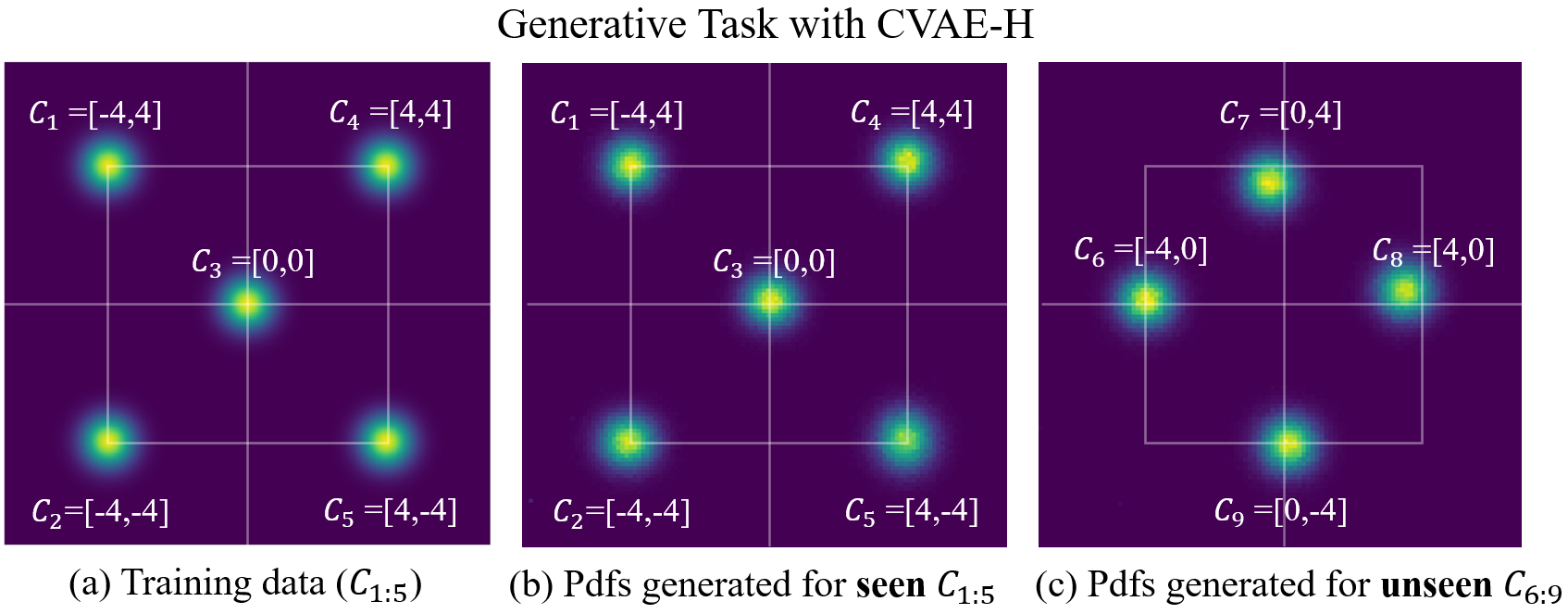}
    \caption{Qualitative results of the Gaussian 2 experiment for the generation task with CVAE-H. (a) data distribution $p(x,y|C_{train})$, b) the data distribution reproduced by CVAE-H $p_{model}(x,y|Z,C_{train})$, and c) CVAE-H's sample generation in \textit{unseen} conditions $p_{model}(x,y|Z,C_{unseen})$.}
    \label{Fig:HCNAF_Gaussians_Generalization_CVAE-H}
\end{figure}

The results of the generative task with CVAE-H are presented in Figure \ref{Fig:HCNAF_Gaussians_Generalization_CVAE-H} and in Table \ref{Table:HCNAF_Gaussians_Generalization_CVAE-H}. Note that the cross-entropy $H(p, p_{model})$ is lower-bounded by $H(p)$ as $H(p, p_{model}) = H(p) + D_{KL}(p||p_{model})$. The differential entropy $H(p)$ of an isotropic bi-variate Gaussian distribution $p(x,y)$ is analytically obtained as $H(p)=0.5\cdot ln(2\pi e(\sigma)^2)^2$. The generative cross-entropy for seen conditions (i.e., $H(p,p_{model}(X_{target}|Z,C_{train}))$) are very close to the target differential entropy. On the other side, the generative cross-entropy for unseen conditions is relatively higher. This quantification agrees with the generated pdfs illustrated in Figure \ref{Fig:HCNAF_Gaussians_Generalization_CVAE-H} that shows CVAE-H's capability to generalize over unseen conditions.

\begin{table}[ht]
\centering
\caption{Differences between the target and \emph{generative} distributions by CVAE-H in terms of cross-entropy and KL divergence for Figure \ref{Fig:HCNAF_Gaussians_Generalization_CVAE-H}.}
\label{Table:HCNAF_Gaussians_Generalization_CVAE-H}
\begin{tabular}{
>{\centering\arraybackslash}p{2.0cm}
>{\centering\arraybackslash}p{1.0cm} 
>{\centering\arraybackslash}p{1.75cm}
>{\centering\arraybackslash}p{1.75cm} }
\toprule
& $p(x, y)$ & \multicolumn{2}{c}{$p_{model}(x,y|Z,C_{i})$} \\
\toprule
$C$ & - &  $C_{i} \in C_{train}$ & $C_{i} \in C_{unseen}$ \\
\midrule
$H(p)$  & 1.452 & - & - \\
\midrule
$H(p,p_{model})$  & - & 1.480 & 2.256 \\
\midrule
$D_{KL}(p||p_{model})$  & - & 0.028 & 0.804 \\
\bottomrule
\end{tabular}
\end{table}

In summary, the Gaussian experiments empirically verified that CVAE-H satisfies all four requirements of the ideal prediction model; probabilistic, multi-modal, context-driven, and generalizable.

\section{Experiments: Multi-agent Forecasting}
\label{sec:evaluation_pred_urban_driving}

In this section, we demonstrate the performance of CVAE-H in urban driving scenarios using PRECOG-Carla dataset \cite{rhinehart2019precog}, which a publicly available dataset created using an open-source simulator called Carla \cite{dosovitskiy2017carla}.

\subsection{The Dataset}
\label{sec:dataset}

PRECOG-Carla dataset includes roughly 76,000 urban driving examples each consists of a 6-second scenario, with up to five road-agents (i.e., simulated human-driven vehicles) driving in urban areas. The first two seconds (i.e., $t={-2:0s}$) are used as inputs to predict the future 4-second (i.e., $t={0:4s}$) trajectories of the road agents. Lidar point clouds of the environment is provided with three overhead lidar channels (i.e., two above ground and one ground level inputs). The goal is to produce accurate predictions of all road-agents using CVAE-H.

The 76,000 clips in the PRECOG-Carla Town01 dataset are split into a training set (80\%), validation set (10\%), and test set (10\%). CVAE-H is trained on the PRECOG-Carla Town01-train dataset and the progress was validated over Town01-val dataset. Town01-test dataset is used to evaluate the performance of the models. It is important to note that \emph{all qualitative and quantitative results} presented in this paper are based on \emph{the test set}. In other words, the models were tested with \emph{unseen} data.

We apply coordinate transformations to obtain \emph{target-centric} coordinates as we found this helps to improve the accuracy of the multi-agent forecasting. Specifically, the coordinates of all give vehicles in the scene are transformed so that $A_{k}$, the target vehicle of the prediction, is positioned at (0,0) at zero heading angle at $t=0$. That is, $X^{A_{k}}_{t=0}:=[0,0]$ and $V^{A_{k}}_{t=0} \cdot [0,1] = 0$. The other road-agents' positions are transformed using the same transformation matrix. As there exists five road-agents in each example, we create 5 instances with each and every one of the road-agents to be located at the center $(0,0)$ at $t=0$. That is, $[X^{A_{\forall k}}_{-2:4s}]^{A_{m}}, m \in {1,2,3,4,5}$, where $[X^{A_{\forall k}}_{-2:4s}]^{A_{m}}$ indicates that $X^{A_{\forall k}}_{-2:4s}$ is transformed in the $A_{m}$-centric coordinates.

\subsection{Hyper-network Design}

The hypernetwork of CVAE-H takes the large condition $C$, processes it, and encodes it into the network parameters for the VAE. Unlike the Gaussian experiments that work with low-dimensional $C$, the condition for the self-driving prediction task is much more high-dimensional. $C$ is defined as $C:=[\Omega, X^{A_{\forall k}}_{-2:0s}]$, where $\Omega \in \mathbb{R}^{3 \times 200 \times 200}$ indicates the 200 by 200 3-channel lidar scans of the environment and $X^{A_{\forall k}}_{-2:0s} \in \mathbb{R}^{5 \times 10 \times 2}$ represents the labelled trajectories of 5 road agents over 10 time-steps. In this regard, a dedicated hypernetwork architecture must be used to effectively extract the spatial and social features of the inputs. Through changes in the hyper-network, CVAE-H can scale up to tackle the high-dimensional multi-agent forecasting problems for autonomous driving. 

\begin{figure}[ht] 
    \centering
    \includegraphics[width=\linewidth]{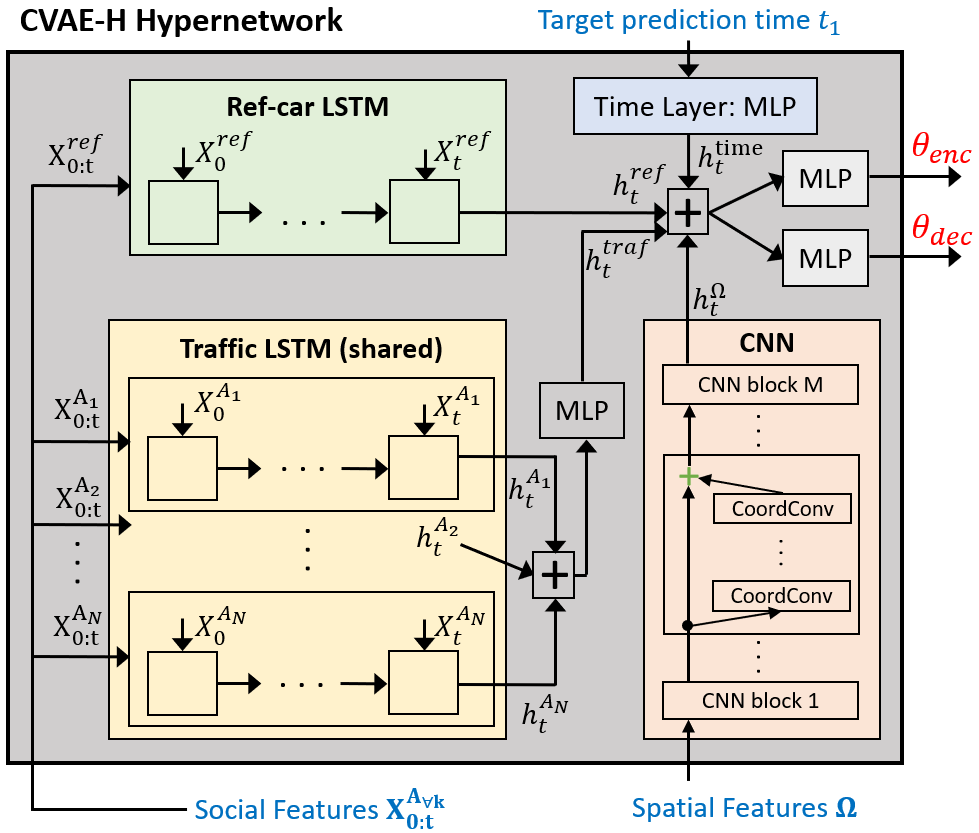}
    \caption{The hypernetwork design for the multi-agent forecasting for autonomous driving. The inputs to the hypernetwork are highlighted in \textcolor{blue}{blue} and the outputs are colored \textcolor{red}{red}.}
    \label{Fig:CVAE-H_hypernetwork}
\end{figure}

Figure \ref{Fig:CVAE-H_hypernetwork} depicts the customized hypernetwork design for the multi-agent forecasting. The hypernetwork takes perception inputs as the aforementioned condition $C$ and outputs a set of network parameters $\theta=[\theta_{enc}, \theta_{dec}]$ that are network parameters for the encoder $Z=f_{enc}(X; \theta_{enc})$ and decoder $X=f_{dec}(Z; \theta_{dec})$. 

Our hypernetwork design was inspired by HCNAF \cite{oh2020hcnaf}. Precisely, we leverage the social, spatial, and time modules of \cite{oh2020hcnaf}. The difference is that our hypernetwork works with \emph{target-centric} coordinates $[X^{A_{\forall k}}_{-2:4s}]^{A_{m}}$ whereas HCNAF works with \emph{AV-centric} coordinate as HCNAF performs an occupancy forecasting that is independently obtained from the road agents. Another difference is that our hypernetwork models network parameters for VAE $\theta_{enc}$ and $\theta_{dec}$, as opposed to the parameters of normalizing flow of HCNAF. 

The details of the hypernetwork design is as follows. As aforementioned, $C$ is formed with spatial features $\Omega$ (i.e., lidar scan) and social features $X^{A_{\forall k}}_{-2:0s}$ (i.e., labelled trajectories of the road agents). Technically, the spatial inputs can come from any sensor outputs (e.g., lidar, camera, or map), but we use lidar data only since that is what the PRECOG-Carla dataset provides. The hyper-network consists of three components: (1) a social module, (2) a spatial module, and 3) a time module. The outputs of the three modules $h^{S}, h^{\Omega}, h^{T}$ are concatenated and fed into two separate 3-layer MLPs with Relu activations, each dedicated to produce network parameters for encoder and decoder, as shown in Figure \ref{Fig:CVAE-H_hypernetwork}. 

The social module is designed using Long Short Term Memory (LSTM) \cite{hochreiter1997LSTM}.
Each LSTM module takes $X^{A_{k}}_{-2:0s} := [x^{A_{k}}_{t}, y^{A_{k}}_{t}]$, the historical states of a road agent in the scene and encodes temporal dependencies and trends among the state parameters. We use two 2-layer LSTM modules; one for encoding the hidden states of road-agents except the target vehicle ($h^{traffic} \in \mathbb{R}^{40}$) and the other for encoding the hidden state of the target vehicle ($h^{ref} \in \mathbb{R}^{24}$). This is because the target is positioned at (0,0) and others somewhere else. 
The resulting output latent vector is the concatenation of the two; $h^{social} = [h^{ref} \oplus h^{traffic}]$.

The spatial module takes in the processed spatial data, which is denoted as $\Omega$, from external perception modules. Our spatial module is based on convolutional neural networks that use residual connections. Recall that the backbone network works with Cartesian (x,y) coordinates space and pixel (image) space. To strengthen the association between the 1D numerical coordinates and 2D image data, we use the coordinate convolution layers \cite{ref:2018_CoordConv} as opposed to regular convolution layers. Overall, the spatial module consists of 4 blocks where each block consists of 5 coordconv layers with residual connections, max-pooling layers, and batch-normalization layers. The output latent variable for spatial feature is $h^{\Omega} \in \mathbb{R}^{96}$.

Lastly, the time module encodes the temporal information into a latent vector $h^{time} \in \mathbb{R}^{10}$ and consists of a 3-layer MLP. The input to the module is the target forecasting time $t_{1} \in \mathbb{R}^{1}$.

\subsection{Qualitative Results}

The following two subsection share the results of the trajectory forecasting experiment. We divide the evaluation into qualitative (visualizations) and quantitative evaluations (accuracy metrics).

\begin{figure}[h] 
    \centering
    \includegraphics[width=0.95\linewidth]{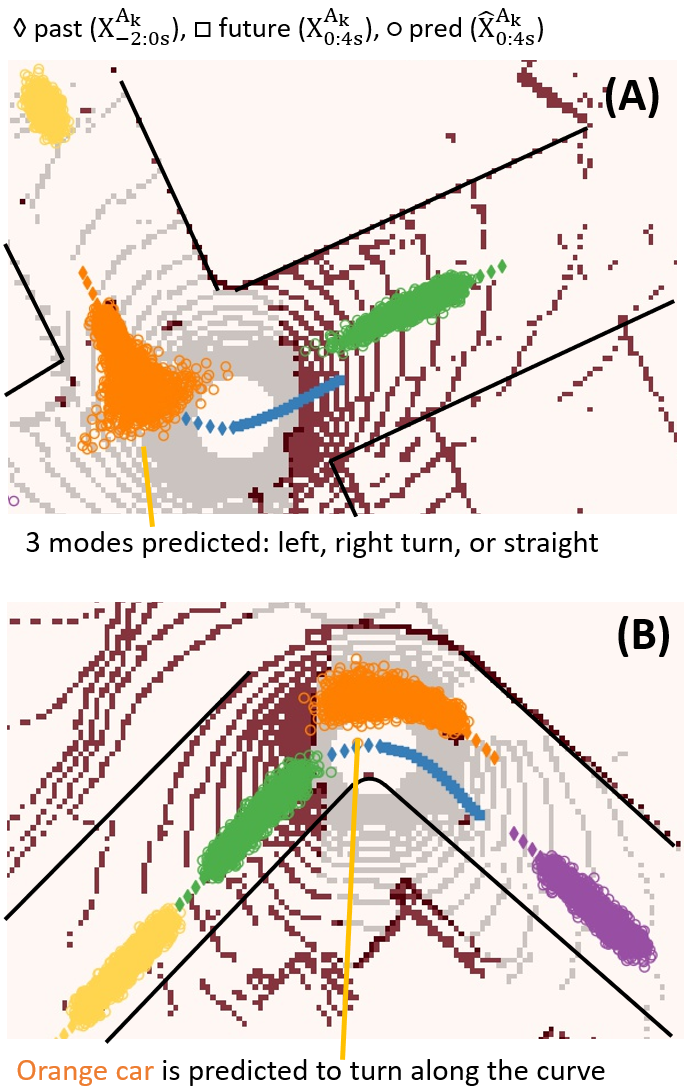}
    \caption{The first set of examples qualitatively shows that CVAE-H predictions are contextual to \emph{spatial} features of the environments (i.e., road topologies) and \emph{multi-modal}.}
    \label{Fig:PRECOG_Trajectory_CVAE-H_spatial}
\end{figure}

\begin{figure}[h] 
    \centering
    \includegraphics[width=0.95\linewidth]{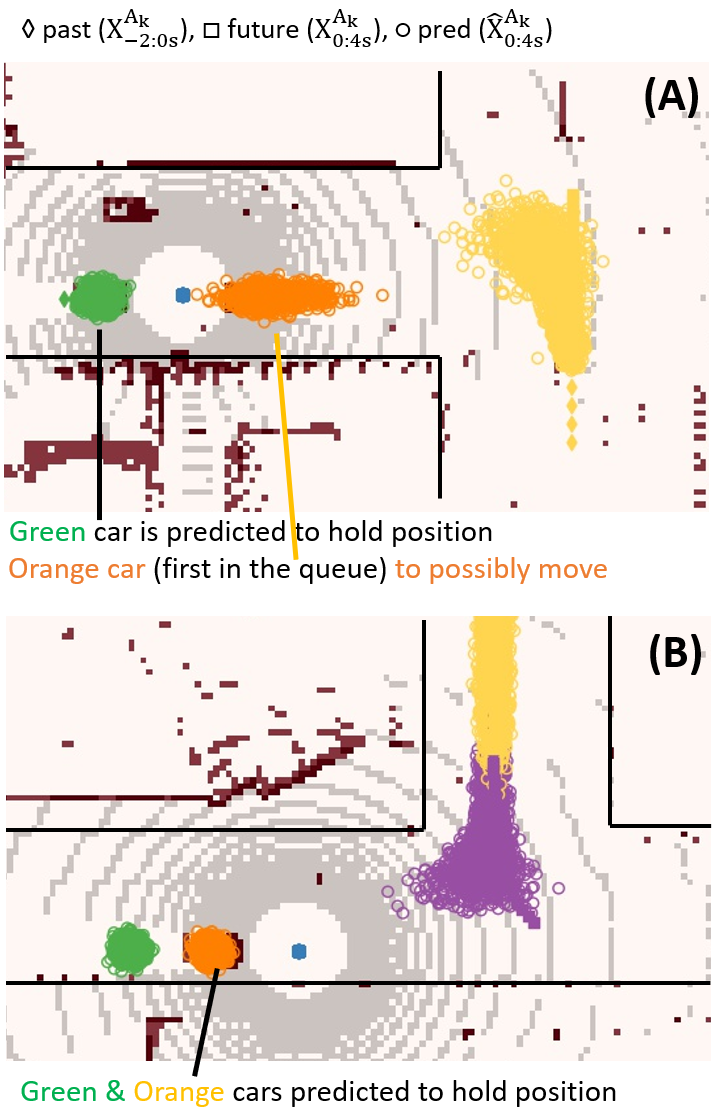}
    \caption{Qualitative demonstrations of \emph{socially contextual} and \emph{multi-modal} CVAE-H outputs.}
    \label{Fig:PRECOG_Trajectory_CVAE-H_social}
\end{figure}

In this subsection, we present the qualitative results in Figure \ref{Fig:PRECOG_Trajectory_CVAE-H_spatial} and \ref{Fig:PRECOG_Trajectory_CVAE-H_social}. In Figure \ref{Fig:PRECOG_Trajectory_CVAE-H_spatial}, we show how CVAE-H outputs are contextual to spatial features of environments. Figure \ref{Fig:PRECOG_Trajectory_CVAE-H_social} illustrates how CVAE-H outputs are contextual to social features.


In each figure, two examples from the PRECOG-Carla Town01 test set are depicted. Each example includes the inputs to the model (i.e., lidar scans of the environment and the 2-second historical states of the road agents), the 4-second label future trajectories, and the outputs of the model (i.e., CVAE-H predictions). We produce 100 sample predictions per car so that the diversity and multi-modality of the outputs are clearly displayed. The estimated road boundaries are overlaid using black lines.

Figure \ref{Fig:PRECOG_Trajectory_CVAE-H_spatial} visualizes that CVAE-H predictions are contextual to the spatial features of the environments. The top figure portrays a 4-way intersection environment where the orange car is entering the intersection, and the green and yellow cars are cruising. A vehicle that approaches a 4-way intersection typically has 3 distinct options (or modes): left-turn, right-turn, or pass-through. The figure suggests that CVAE-H captured all three modes. For green and yellow cars, CVAE-H predicted them to cruise, as reflected in the predictions that progressed towards the direction of travels.

The bottom plot of Figure \ref{Fig:PRECOG_Trajectory_CVAE-H_spatial} depicts a curved-road environment with five cars in the scene. The orange car is driving in the corner of the road, and the green, yellow, and purple cars are either approaching or leaving the curved road. The output prediction for the orange car shows that CVAE-H accounts for the spatial constraint of the curve and makes accurate predictions along the curvature of the road.

CVAE-H considers the social contexts of the environment as displayed in Figure \ref{Fig:PRECOG_Trajectory_CVAE-H_social}. First, let us pay attention to the orange and green cars in the figure. In the top plot, the orange car is the first in the queue, whereas the orange car is not in the bottom plot. CVAE-H understands this social context and accordingly predicts that the orange car in the top environment may move in the x-direction, as shown in the elongated uncertainty in the x-direction, whereas the orange car in the bottom is most likely to stay idle. Indeed, this is how a queue is cleared. Secondly, the green car in both top and bottom environments are located last in the queue; CVAE-H understands that there are two other vehicles in front and accordingly predicts that the green car would not move. Lastly, CVAE-H captures the multi-modality of the yellow car in the top environment and the purple car in the bottom environment. 

We emphasize that CVAE-H not only outputs predictions, but also the probability associated with them. Although the certainty of the prediction is not used in this work as it is out of scope, the access to the probability is essential for planning tasks. The prediction probability is obtained through the final output layer of CVAE-H, which is a Gaussian mixture. In this regard, CVAE-H is \emph{probabilitic}. As aforementioned, all qualitative examples presented in this section were produced using a \emph{single} CVAE-H model. This empirically shows that CVAE-H is \emph{general}. In this sense, CVAE-H possesses all four attributes of the ideal prediction model.

\subsection{Quantitative Results}

For quantitative evaluations, we use the minimum mean squared deviation of sampled trajectories from the ground-truth (\emph{minMSD}). MinMSD is a measure of diversity or multi-modality of the model outputs. It is often used for evaluating self-driving prediction models, in particular, generative prediction models that construct probabilistic models with sampling capability \cite{rhinehart2019precog, rhinehart2018r2p2, tang2019MFP, chai2019multipath}. The definition of MinMSD is as follows.

\begin{equation}
\label{Eq:definition_minMSD}
\begin{aligned}
\hat{m}_{K} = min_{k \in {1..K}} [||X^{k}-X^{gt}||^2/T],
\end{aligned}
\end{equation}

where $K$, $X^{k}$, $X^{gt}$, and $T$ corresponds to the number of trajectory samples, k-th sample trajectory, ground-truth trajectory, and the prediction horizon. While there exist similar metrics like minimum average displacement error (minADE) and minimum final displace error (minFDE) \cite{tang2019MFP, chai2019multipath, phan2020covernet}, we report minMSD to compare our models against the available benchmarks.

Table \ref{Table:results_PRECOG_Carla_minMSD} presents the evaluation results against available benchmarks. Our model outperforms most of these prediction models \cite{lee2017desire, gupta2018socialGAN, rhinehart2018r2p2, rhinehart2019precog, chai2019multipath}, except MFP \cite{tang2019MFP}. Although MFP achieved the best performance, MFP is a deterministic model that outputs a fixed number of deterministic modes (i.e., trajectories), compared to other methods including ours that outputs a non-deterministic trajectories.

\begin{table}[h]
\caption{Quantitative evaluation of the proposed prediction models using MinMSD with K=12 on PRECOG-Carla town01 testset, averaged over all 5 cars. Lower is better.}
\label{Table:results_PRECOG_Carla_minMSD}
\begin{tabular}{ 
p{2.1cm} 
p{3.5cm} 
p{1.4cm}
}
\toprule
\textbf{Method} & \textbf{minMSD($m^2$)}, K=12 \\
\toprule
GAN-based & SocialGAN \cite{gupta2018socialGAN}  & 1.464 \\
\midrule
\multirow{2}{*}{Flow-based} & R2P2 \cite{rhinehart2018r2p2}   & 0.843 \\
\cmidrule{2-3}
& PRECOG-ESP \cite{rhinehart2019precog}   & 0.716 \\
\midrule
GMM-based & MultiPath \cite{chai2019multipath}  & 0.680 \\
\midrule
AE-based  & \multirow{2}{*}{MFP \cite{tang2019MFP}}   & \multirow{2}{*}{\textbf{0.279}} \\
(deterministic) & & \\
\midrule
\multirow{2}{*}{VAE-based} & DESIRE \cite{lee2017desire}   & 2.599 \\
\cmidrule{2-3}
& CVAE-H (ours)         & \textbf{0.312} \\
\bottomrule
\end{tabular}
\end{table}

While Table \ref{Table:results_PRECOG_Carla_minMSD} presented minMSD averaged over all 5 cars, Table \ref{Table:results_PRECOG_Carla_minMSD_detailed} shares the detailed minMSD numbers per road agent in the scene against available benchmarks. 

\begin{table}[h]
\caption{The detailed minMSD results for individual agents.}
\label{Table:results_PRECOG_Carla_minMSD_detailed}
\begin{tabular}{ 
p{1.7cm} 
p{0.8cm}
p{0.8cm}
p{0.8cm}
p{0.8cm}
p{0.8cm}
}
\toprule
\textbf{Method} & Car 1 & Car 2 & Car 3 & Car 4 & Car 5 \\
\toprule
DESIRE \cite{lee2017desire} & 2.621 & 2.422 & 2.710 & 2.969 & 2.391 \\
\midrule
PRECOG-ESP \cite{rhinehart2019precog} & 0.340 & 0.759 & 0.809 & 0.851 & 0.828 \\
\midrule
CVAE-H (ours) & \textbf{0.151} & \textbf{0.346} & \textbf{0.304} & \textbf{0.355} & \textbf{0.404} \\
\bottomrule
\end{tabular}
\end{table}

\vspace{-8pt}
\section{Conclusion}

We present CVAE-H, a novel variational autoencoder conditionalized using a hypernetwork, and demonstrate that CVAE-H is an ideal model for the high-dimensional multi-agent forecasting problem of self-driving. Unlike the traditional embedding approach to make a VAE conditional, CVAE-H offers an unique way of conditioning a VAE using the hypernetwork architecture. This facilitates the informational flow of the conditions to the VAE network across all layers and thus allows the VAE to scale up to large conditional inputs of the prediction problem. We first evaluated CVAE-H in simple generative tasks where we demonstrated the data generation and generalization capabilities of the proposed model. Then, we scaled up the hypernetwork to work with a high-dimensional trajectory forecasting task for autonomous driving. We showed that CVAE-H is probabilistic, multi-modal, context-driven, general, and that it achieves a comparable performance to the state-of-the-art prediction model.

{\small
\bibliography{egbib}
}

\clearpage

\beginsupplement

\section{More Qualitative Results of the Multi-agent Forecasting}
\label{sec:more_qualitative_results}

In this section, we present more qualitative results of the multi-agent forecasting. All results are based on the Town01 test set of the PRECOG-Carla dataset.

\begin{figure*}[ht] 
    \centering
    \includegraphics[width=\linewidth]{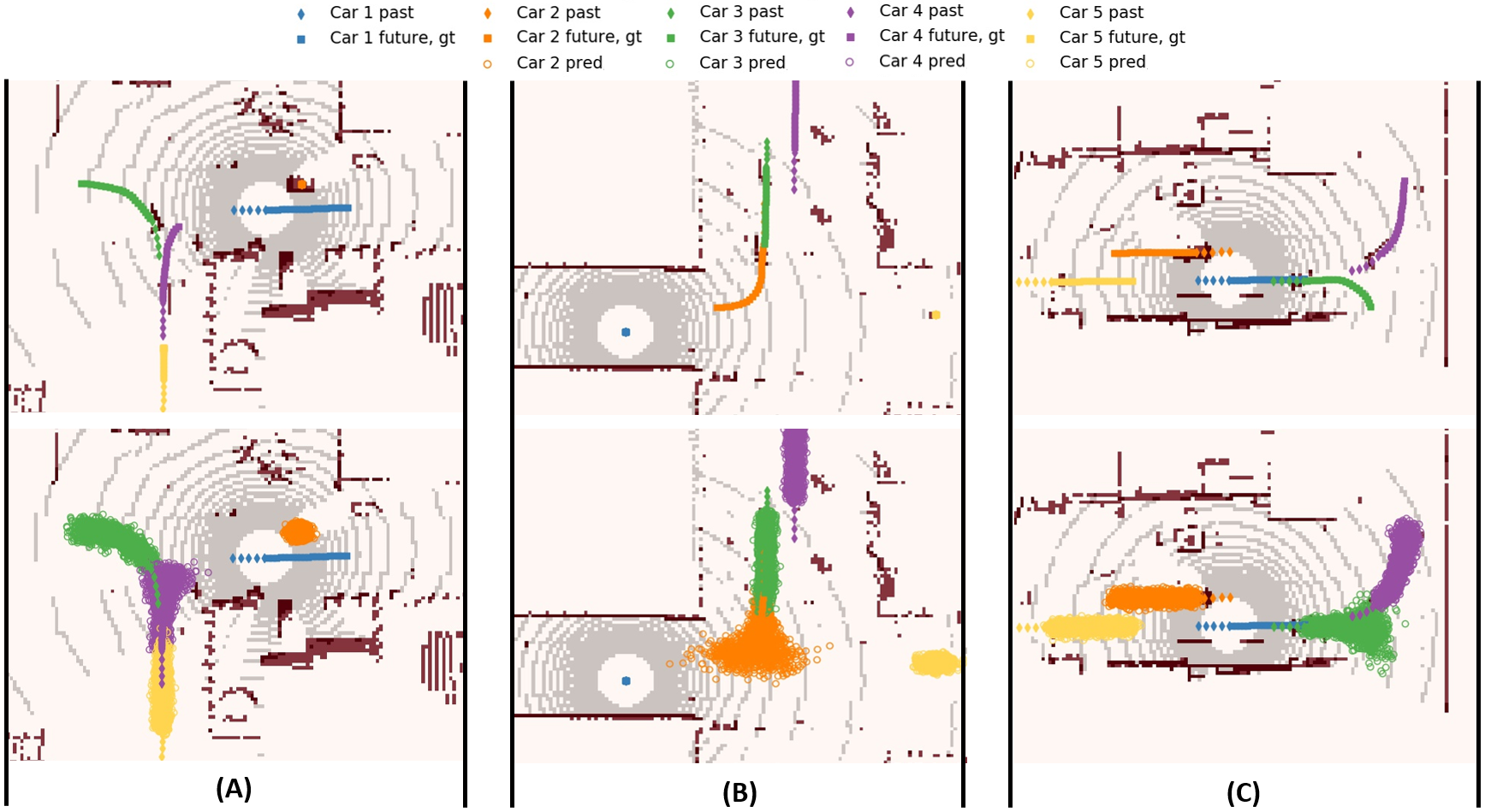}
    \caption{More qualitative examples. Each example consists of a top and bottom plot. A top plot depicts the input to the prediction model as well as the ground-truth future positions of the road agents. The input consists of a lidar scan of the environment $\Omega$ and 2-second history $X^{A_{\forall k}}_{-2:0s}$ (depicted as $\diamond$). The 4-second ground-truth future positions $X^{A_{\forall k}}_{0:4s}$ are depicted using $\square$. A bottom plot depicts $\hat{X}^{A_{\forall k}}_{0:4s}$, the outputs of the prediction model overlaid over future time-steps, $t=0:4s$. In other words, a bottom plot describes how the predictions progress over time.}
    \label{Fig:More_qualitative_results_1}
\end{figure*}

Figure \ref{Fig:More_qualitative_results_1} illustrates 3 examples from the test set. The first example (A) depicts a three-way environment where the purple and yellow cars are entering the intersection, the green car is making a left-turn, and the orange car is waiting. For the purple car, CVAE-H predicted it to have two modes; either turning left or right. For the green car, the proposed prediction model predicted it to finish a left-turn with the correct turning curvature, which suggests that CVAE-H took account for the spatial features of the environment. CVAE-H accurately predicted the orange car to hold the position and the yellow car to cruise. 

The second example (B) describes a 4-way intersection environment where the green and orange cars are entering the intersection, the purple car leaving the intersection, and the yellow car is waiting. Among the four road agents, the orange car is the most multi-modal as it could possibly turn left, right, or pass through the intersection. The predictions from CVAE-H successfully captured all three modes. CVAE-H predicted the green and purple cars to cruise. The yellow car is mostly predicted to hold the position, but a small possibility to start moving is reflected as the uncertainties in the x-axis.

The third example (C) is about another 3-way intersection environment where the green is entering the intersection, the purple car is on a left-turn, and the orange and yellow cars are cruising. For the green car which is about to enter the intersection, CVAE-H predicted that it could make either a left or right turn. For the purple car which is already on a left-turn, it was predicted to finish the turn toward the end of the prediction horizon. CVAE-H predicted the yellow and orange cars to continue to cruise.

\begin{figure*}[ht] 
    \centering
    \includegraphics[width=\linewidth]{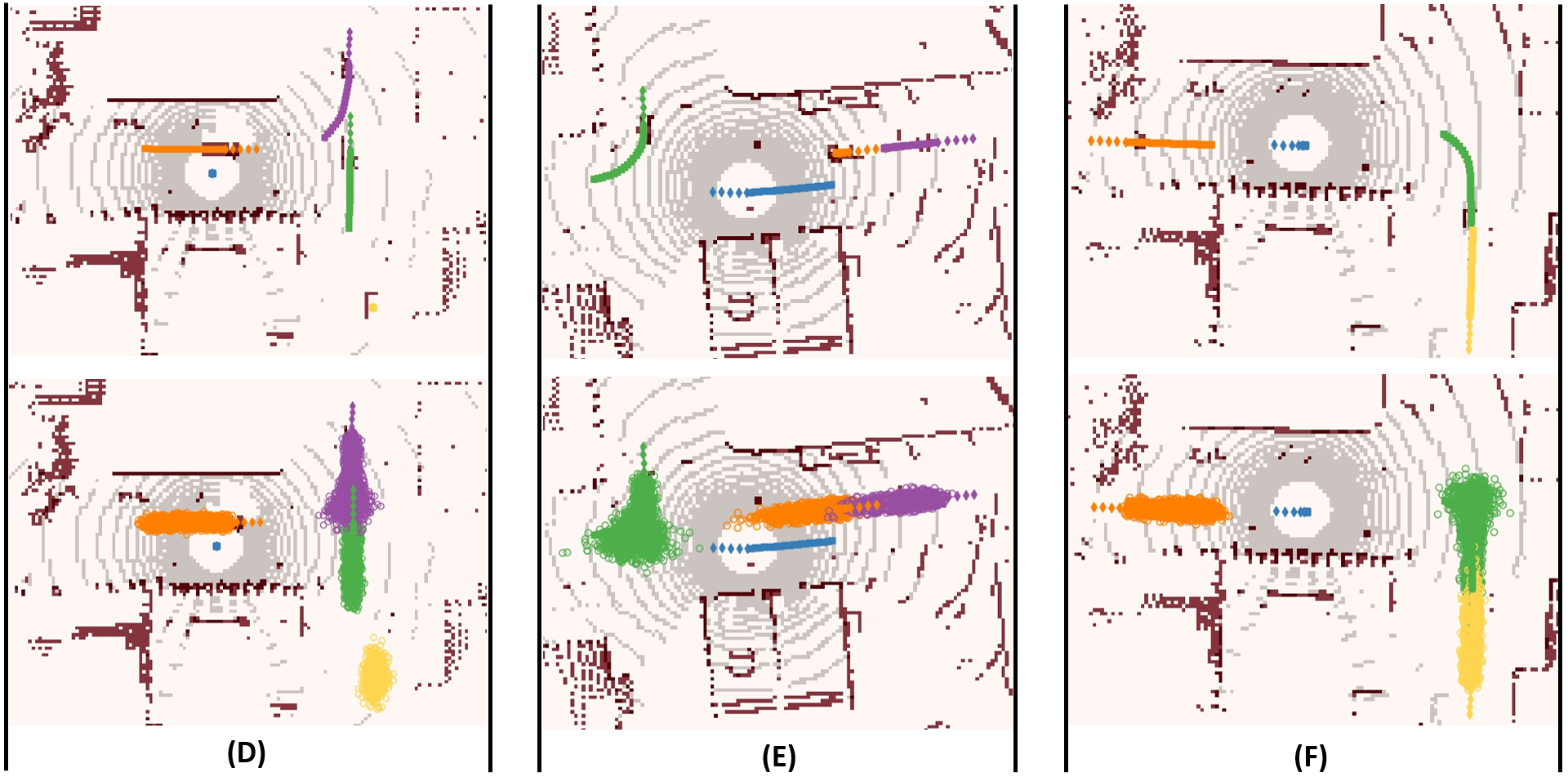}
    \caption{Another 3 examples (D-F) of the multi-agent forecasting with CVAE-H. The explanations about each example are provided in the body of the appendix.}
    \label{Fig:More_qualitative_results_2}
\end{figure*}

Figure \ref{Fig:More_qualitative_results_2} illustrates another 3 examples from the test set. Example (D) describes a 4-way intersection environment where the purple car is entering the intersection, the green car is in the middle of the intersection-passing, the orange car is leaving the intersection, and the yellow is waiting for its turn. The purple car could possibly either turn left, right, or pass through the intersection. CVAE-H captured all three modes successfully. For the green car, CVAE-H predicted it to continue to pass through the intersection, as it was already on its way through the intersection. The proposed prediction model predicted the orange car to cruise and the yellow car to hold its position. 

Example (E) illustrates a 4-way intersection where the orange and purple cars are approaching the intersection, and the green car is entering the intersection. As the green car just started to make a right-turn, the prediction weights more towards the right-turn and pass-through modes, but still captured a small possibility of the left-turn mode. For the orange and purple cars, CVAE-H predicted them to continue to cruise. 

Example (F) depicts another 4-way intersection where the green and yellow cars are approaching the intersection, and the orange car is cruising behind the blue reference car. The green car is predicted to have three modes; straight, left-turn, and right-turn. The orange and yellow cars are predicted to cruise.

\end{document}